\documentclass{article}
   \PassOptionsToPackage{numbers, compress}{natbib}

     \usepackage[final]{neurips_2019}
\usepackage{cite}
\usepackage{natbib}
\usepackage[utf8]{inputenc} 
\usepackage[T1]{fontenc}    
\usepackage{hyperref}       
\usepackage{url}            
\usepackage{booktabs}       
\usepackage{amsfonts}       
\usepackage{nicefrac}       
\usepackage{microtype}      
\usepackage{cite}
\usepackage{amsmath,amssymb,amsfonts}
\usepackage{graphicx}
\usepackage{textcomp}
\usepackage{caption}
\usepackage{comment}
\usepackage{setspace}
\usepackage{mathtools}
\usepackage{float}
\usepackage{graphics}
\usepackage{titlesec}
\usepackage{subfig}
\usepackage{graphicx}
\usepackage{xcolor}
\usepackage{titlesec}
\usepackage{graphicx}
\usepackage{mathtools}
\usepackage{siunitx}
\usepackage{soul}
\usepackage{balance}
\usepackage{algpseudocode}
\usepackage{float}
\usepackage{cite}
\usepackage[font=footnotesize]{caption}
\usepackage{tabularx}
\usepackage{array}
\usepackage[english]{babel}
\usepackage{xcolor}
\usepackage{tikz}
\usepackage{blindtext}
\usepackage{eucal}
\usepackage{soul}
\usepackage{enumerate}
\usepackage[ruled,vlined]{algorithm2e}
\usepackage{algpseudocode}
\title{Supported-BinaryNet: Bitcell Array-based Weight Supports for Dynamic Accuracy-Latency Trade-offs in SRAM-based Binarized Neural Network}

\author{%
  Shamma ~Nasrin\\
  University of Illinois at Chicago\\
  Chicago, IL 60607 \\
  \texttt{snasri2@uic.edu} \\
  \And
  Srikanth Ramakrishna \\
  University of Illinois at Chicago \\
  Chicago, IL 60607 \\
  \texttt{sramak7@uic.edu } \\
  \AND
  Theja Tulabandhula \\
  University of Illinois at Chicago \\
 Chicago, IL 60607 \\
  \texttt{theja@uic.edu} \\
  \And
 Amit Ranjan Trivedi \\
 University of Illinois at Chicago \\
  Chicago, IL 60607 \\
  \texttt{amitrt@uic.edu } \\
}

\begin{document}
\maketitle
\begin{abstract}
In this work, we introduce bitcell array-based support parameters to improve the prediction accuracy of SRAM-based binarized neural network (SRAM-BNN). Our approach enhances the training weight space of SRAM-BNN while requiring minimal overheads to a typical design. More flexibility of the weight space leads to higher prediction accuracy in our design. We adapt row digital-to-analog (DAC) converter, and computing flow in SRAM-BNN for bitcell array-based weight supports. Using the discussed interventions, our scheme also allows a dynamic trade-off of accuracy against latency to address dynamic latency constraints in typical real-time applications. We specifically discuss results on two training cases: (i) learning of support parameters on a pre-trained BNN and (ii) simultaneous learning of supports and weight binarization. In the former case, our approach reduces classification error in MNIST by 35.71\% (error rate decreases from 1.4\% to 0.91\%). In the latter case, the error is reduced by 27.65\% (error rate decreases from 1.4\% to 1.13\%). To reduce the power overheads, we propose a dynamic drop out a part of the support parameters. Our architecture can drop out 52\% of the bitcell array-based support parameters without losing accuracy. We also characterize our design under varying degrees of process variability in the transistors.
\end{abstract}

\section{Introduction}
Machine learning (ML) algorithms use growing volume and variety of data, faster computing power, and efficient storage for highly accurate predictions and decision-making in complex computing problems. Consequently, ML is becoming an integral component of computational imaging, speech processing, spam filtering, etc. While the earlier ML applications were generally confined to static models, in the second generation, ML finds application in real-time problems like autonomous vehicles and internet-of-things. Real-time applications require dynamic decision-making based on evolving inputs; thus, require ML models to predict with the least latency. The predictive robustness of ML models also becomes critical since actions are taken immediately based on the predictions.

\begin{figure}[]
        \centering
        \subfloat[]{\includegraphics[width=0.35\linewidth]{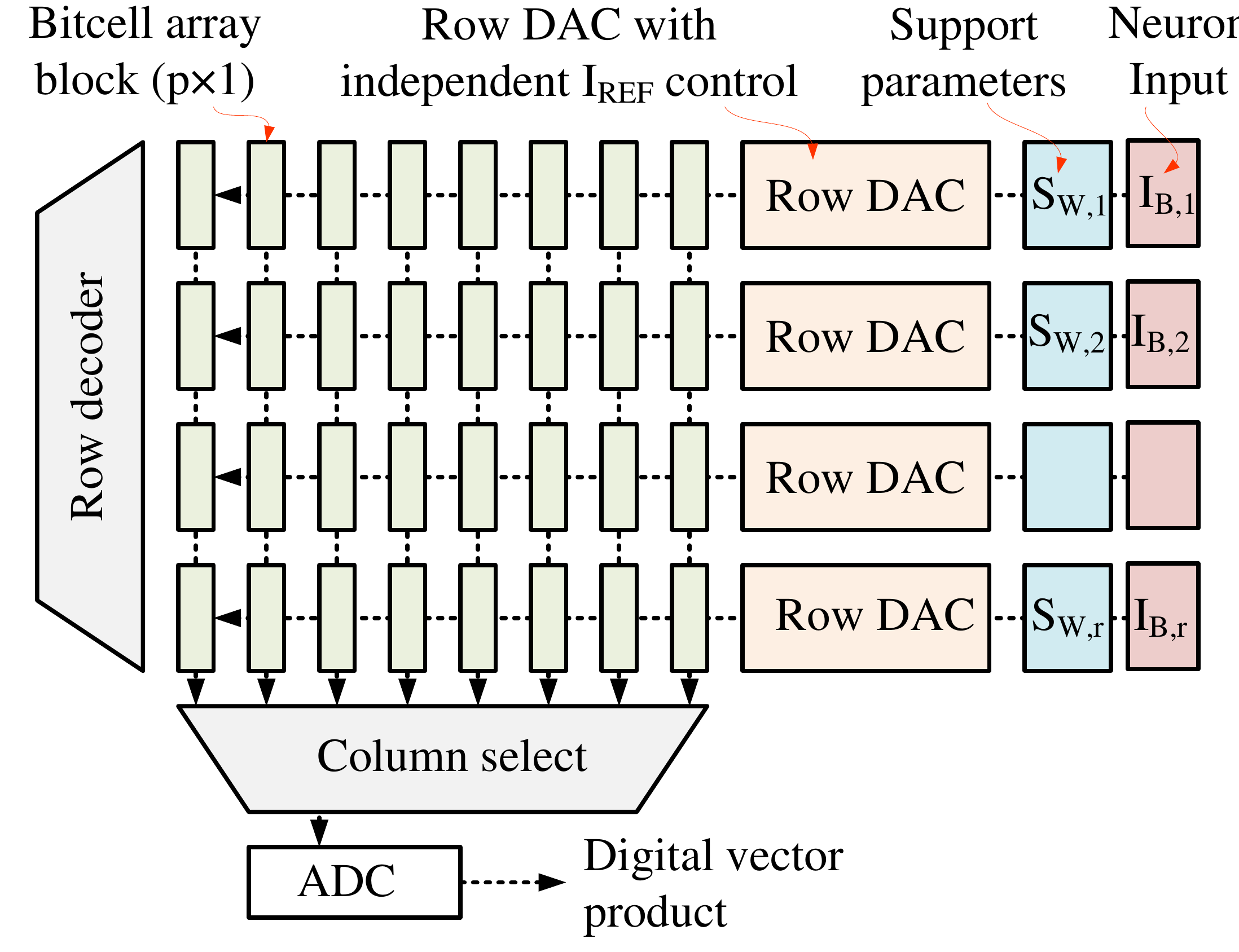}}
        \subfloat[]{\includegraphics[width=0.49\linewidth]{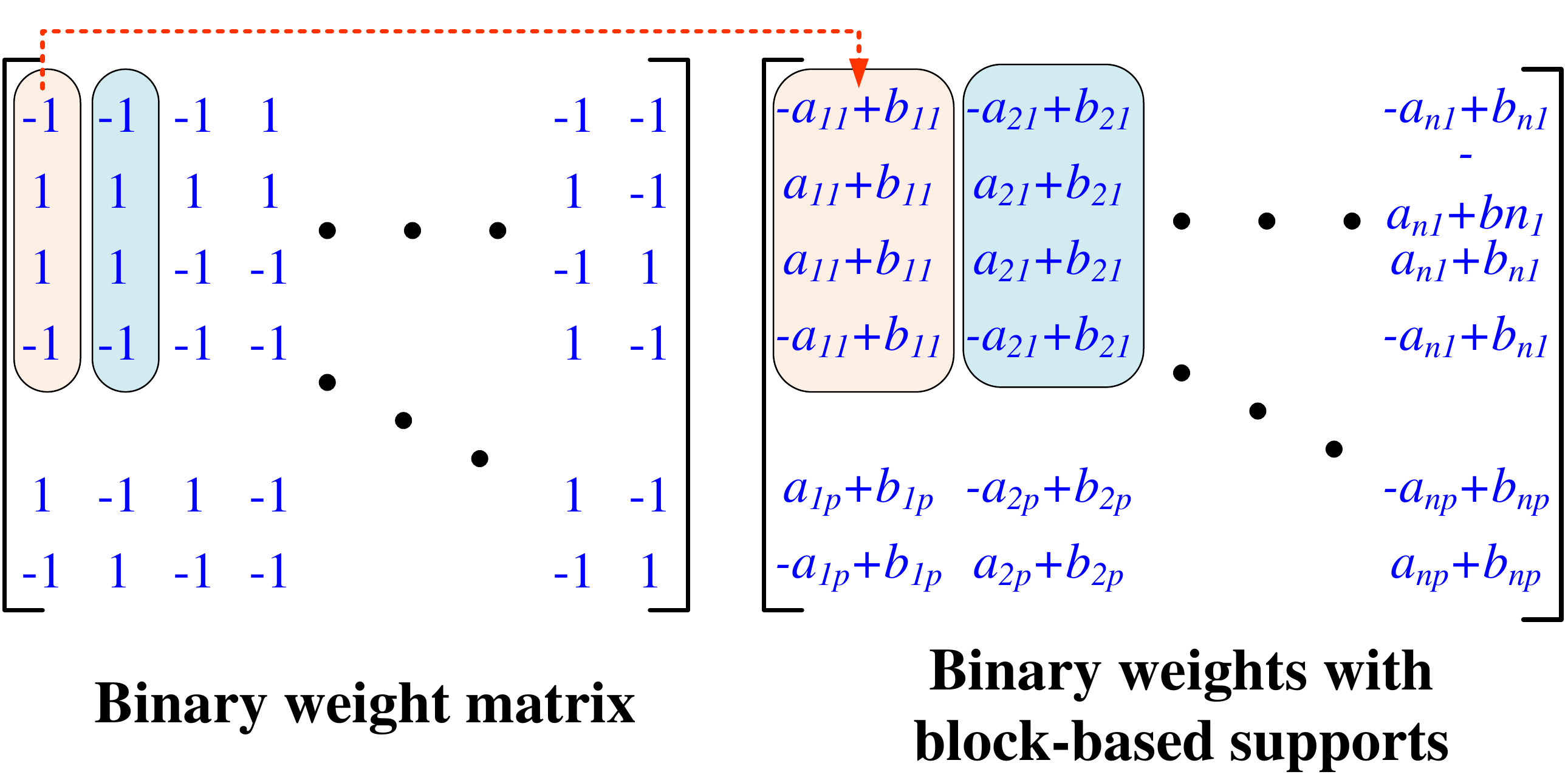}}
        
         \caption{(a) SRAM architecture for bitcell array-based support parameters. Support parameters to enhance weight space of binarized neural networks (BNNs) are stored in the buffer of row digital-to-analog converter (DAC). Row DAC modulates bias-current to implement support parameter-based weight scaling in each bitcell array block. (b) Introducing bitcell array-based support parameters in binarized weight matrix. Here, each block of binary weight matrices is mapped to $aw+b$. $a$ and $b$ are learnable support parameters. } 
         \label{fig:SRAMB}
\vspace{-2em} 
\end{figure}

\begin{figure}[]
   \centering
   \centering
    \subfloat[]{\includegraphics[width=0.19\linewidth]{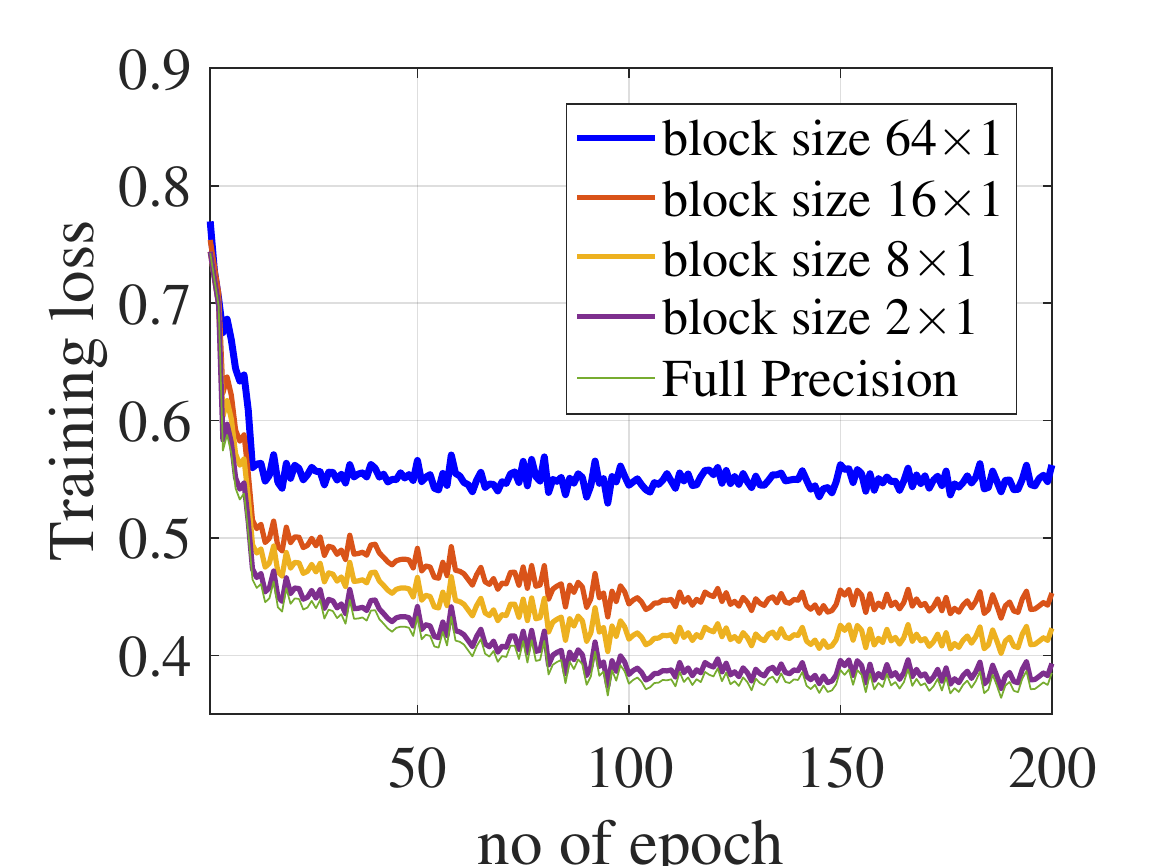}}
    \subfloat[]{\includegraphics[width=0.19\linewidth]{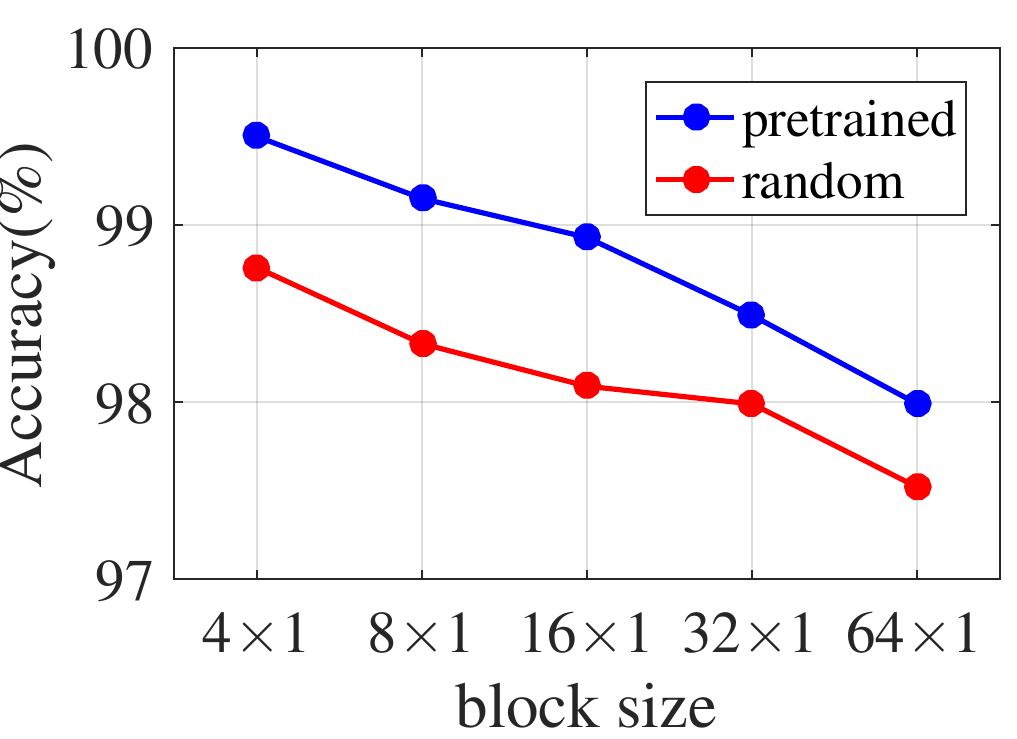}}
    \subfloat[]{\includegraphics[width=0.19\linewidth]{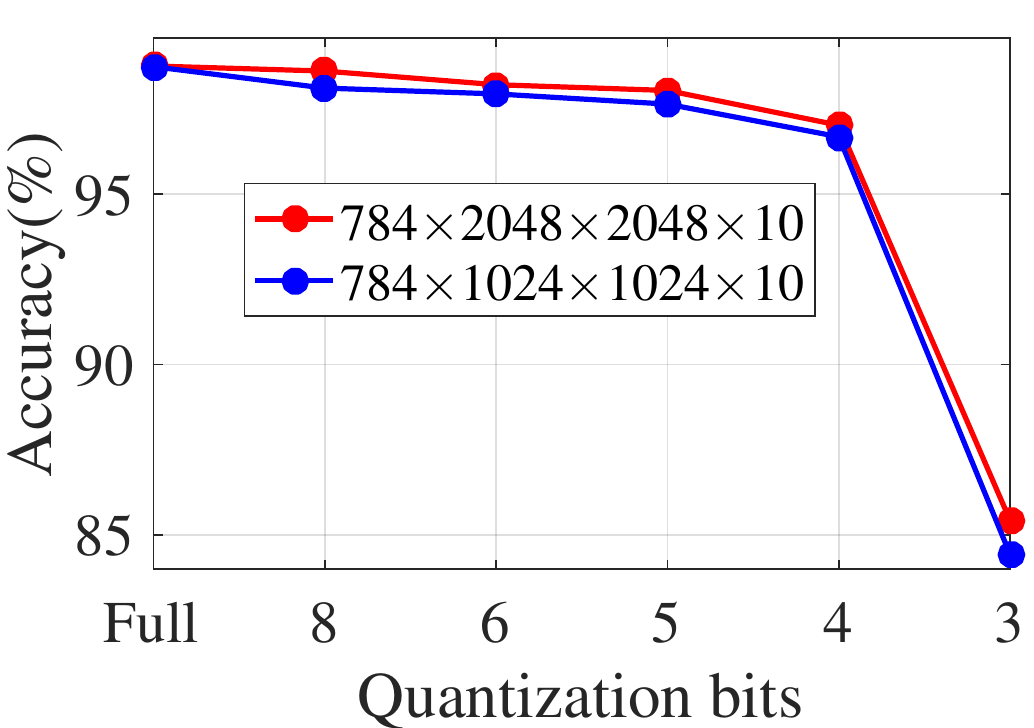}}
    \subfloat[]{\includegraphics[width=0.17\linewidth]{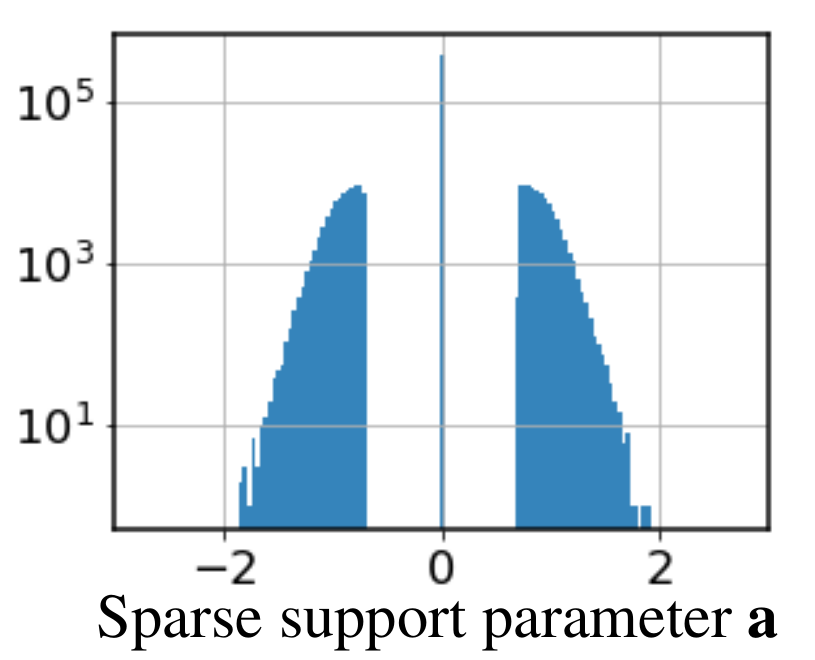}}
      \subfloat[]{\includegraphics[width=0.19\linewidth]{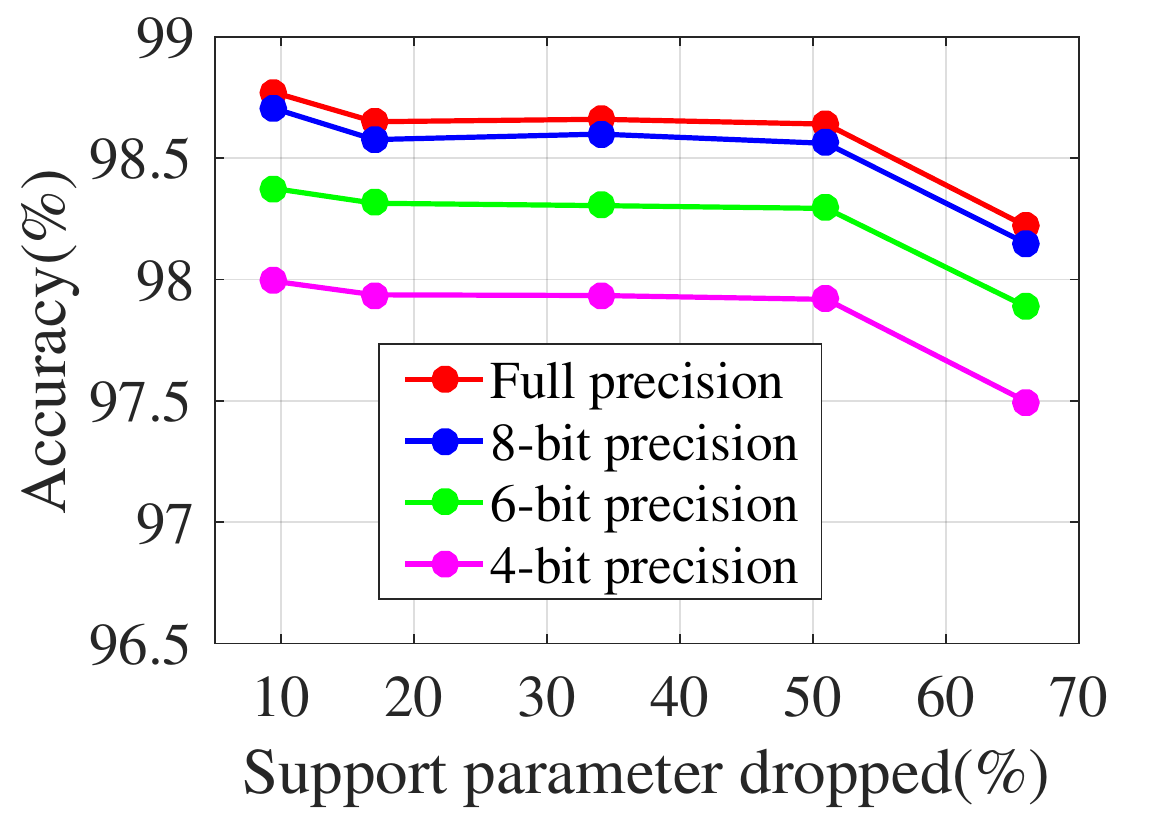}}
    
    \caption{(a) Learning curve of fully connected neural network for MNIST training data set. (b) Accuracy on MNIST by performing support optimization on weights \{-1,+1\} to \{b,a\} for various block sizes. We show results for two cases: (1) initialized with pre-trained network and (2) initialized randomly.(c) Accuracy with different  bit precision of the support parameters. (d) Sparse support parameter $a$ of the 1st layer of the fully connected layer for 52\% support parameter drop. (e) Accuracy on MNIST data-set for varying percentage of support parameter drop with different bit precision of $\epsilon$.}
    \label{fig:acc}
    
\end{figure}
For any ML algorithm, the fundamental computational operations are multiplication and accumulation, which create the energy bottleneck while designing a hardware accelerator. Notably, the number of weights in even a moderate size network for real-world applications can be hundreds to tens of thousands. In such networks, a typical von Neumann platform incurs high traffic to read weights from the memories and to write back neuron output and partial sums. To overcome these limitations, near-memory data processing has been explored where logic units are embedded closer to memories to minimize communication path length. [1] and [2] introduce advanced memory technology like embedded DRAM (eDRAM) in near-memory data processing to reduce the access energy. 

Although techniques such as near-memory data processing and inclusion of advanced memory architecture such as eDRAM and 3-D memories minimize the data transmission demand between memory and logic, the eventual solution is to seamlessly merge logic with memory. In-SRAM neural networks have been discussed in [6] to follow such in-memory computing. In these works, the same SRAM array storing the network weights can also be used to locally compute the vector product of neuron input and weight matrices. However, a key challenge in the current designs is that the network weights must be binarized [7]. Meanwhile, weight binarization leads to higher inaccuracy by limiting the flexibility of weight space. Although operation with multi-bit precision weights is possible for in-SRAM neural networks, the complexity and power of the implementation increase considerably; thereby, quickly decreasing the benefits of in-memory computations. 
 
In this paper, we propose a novel architecture of SRAM-BNN with bitcell array-based support parameters (Fig.\ref{fig:SRAMB}(a)). In this scheme, we partition the binary weight matrix into blocks where the partial-sum output of each block is augmented with support parameters. Weight partitioning is considerate of the physical design and mapping to SRAM, thereby requires minimum interventions to introduce support parameters for each block while maximizing the flexibility of training weight-space. Bitcell array-based supports are implemented by modifying the design of digital-to-analog converter (DAC) at SRAM rows. We also discuss a novel control flow for SRAM-BNN where energy and accuracy can be traded-off dynamically. For MNIST dataset, the architecture achieves 99.24\% prediction accuracy with a pre-trained initial network and 98.87\% accuracy with a randomly initialized network, which is higher than the previous works [5].

Sec. II introduces support vectors in SRAM-BNN. Sec. III implements SRAM-BNN with support vectors. Sec. IV discusses the simulation results. Sec. V concludes.
\section{SRAM Bitcell Array-based Support Vectors}

(Fig.\ref{fig:SRAMB}(b)) shows the overview of bitcell array-based support parameters in a binarized weight matrix. Consider a fully-connected weight matrix of dimensions $m\times n$. We segment the matrix so that each column of the matrix is divided into smaller blocks. For a block size $p$, there will be $m/p$ sub-blocks for each column. Support parameters [$a_{ij}, b_{ij}$] are introduced for each block $B_{ij}$. Using the support parameters, weight values $w \in [-1, 1] $ of a block $B_{ij}$ are mapped to new values such that $w \gets a_{ij}\times w + b_{ij}.$ Therefore, using the support parameters, the weight space of BNN is enhanced. At smaller block size, more parameters are introduced, thereby leading to more flexibility in the weight space. Also, note that at $p = 2$, the weight space becomes equivalent to the full-precision case.    

Typical SRAM-BNN operates the weight matrix column by column and breaks a larger matrix into multiple SRAM arrays. Notably, the above support vectors are considerate to such physical implementation of weight matrix on SRAM array -- a feature that we will leverage in the later sections to retain the hardware simplicity while increasing the flexibility of training weight space. The algorithm 1 describes the approach to get the optimized support vector parameters. In the algorithm, we describe training the network both from a given binary weight matrix and from a random weight initialization. 

 \begin{algorithm}[]
 \caption{Algorithm for support learning in SRAM-BNN with random weight initialization. $C$ is the cost function and $L$ is the number of layers. Dimensions of sub-block and weight matrix are $p\times 1$ and $m\times n$, respectively, and $r$ is the total number of sub-blocks. In this algorithm  $\mathcal{F}$(.) is the activation function. The function Binarize() shows how we binarize the weight matrix [5]. Update() is the function to update the parameters.  Here, We use the ADAM's update rule [6]}
 \textbf{Requirement:}  a mini batch of inputs and targets $(X, T)$, previous weights $W$, Previous support parameters $A,B$,  and previous learning rate $\eta$.\\
\textbf{Given:} Input to the layer \textbf{X} $\in$ $\mathbb{R}^{m \times 1}$,\\

\textbf{{1.1\hspace{.5em} \textit{Forward Propagation}}}\\
 \hspace{1.5em}\textbf{For} layer $k$=1 to $L$ \textbf{Do},\\
\hspace{2.0em}a. $W^b \leftarrow Binarize(W)$ \\
\hspace{2.0em}b. Compute support vector $s_{k,j} \leftarrow a_{i,j}(\textbf{w}^{b})+b_{i}$ ,
\\
\hspace{2.0em}i=1,..,r,    \textbf{s},\textbf{w} $\in$ $\mathbb{R}^{p \times 1}$
\\

\hspace{2.0em}c. Compute z$_{ij}$=$\mathcal{F}$($\sum_{k=1}^{m} $ x$_{ik}$$ S_{kj}$), i=1,...,d, j=1,...,n \\
 \hspace{1.5em}\textbf{end For}\\
\textbf{{1.2 \hspace{.5em} \textit{Backward Propagation}}}\\
\hspace{1.5em} \textbf{For }layer k=L to 1 \textbf{Do},
\\
\hspace{2.0em} Compute gradients $\frac{\partial C}{\partial \textbf{a}_{k}}$,$\frac{\partial C}{\partial \textbf{b}_{k}}$ and $\frac{\partial C}{\partial \textbf{W}_{k}}$ \\
\hspace{1.5em}\textbf{end For}\\
\textbf{{1.3 \hspace{.5em} \textit{Updating the parameters}}}\\
\hspace{1.5em}\textbf{For} layer $k$=1 to $L$, \textbf{Do}\\
\hspace{2.0em}a. $w^{k+1} \leftarrow Update (W)$,
\hspace{2.0em}b. $a^{k+1} \leftarrow Update (a)$,
\hspace{2.0em}c. $b^{k+1} \leftarrow Update (b)$\\
\hspace{1.5em}\textbf{end For}
\end{algorithm}

    

We tested our algorithm with the benchmark MNIST handwritten data set using a 4-layer fully-connected network. We initialized the support parameters $a_{ij},b_{ij}$ randomly to simulate bit-array support-based BNN.  
Fig. \ref{fig:acc}(a) shows the learning curve of the four-layer network for varying block sizes. Training loss for block size two is almost similar to the full precision network (as expected), and loss increases with the increasing block size. Fig. \ref{fig:acc}(b) shows the accuracy results on MNIST data set with support vector optimization on weights for random initialization and initialization with a pre-trained network for different block sizes. Later we will discuss that lower precision of support parameters is helpful in reducing power dissipation. Therefore, Fig. \ref{fig:acc}(c) depicts the impact of quantization of the support parameters on the accuracy of the network. In Fig. \ref{fig:acc}(c), the inference accuracy is tolerant up to four-bit precision in the parameters. In Fig. \ref{fig:acc}(d), zero-centered single mode distribution of support parameters, especially `a,' also allows support parameter-based sparsification of BNN weight matrix. Here, if $|a_{ij}| < \epsilon$ than $a_{ij}$ is approximated as zero. If $a_{ij}$ is zero, the corresponding weight matrix block need not be processed against input. Fig. \ref{fig:acc}(e) shows the accuracy results for varying $\epsilon$ and different bit precision of `a.' Note that up to 52\% support parameters can be dropped without a considerable loss in accuracy. A key benefit of support parameter-based pruning is that it allows an easier implementation of dynamic energy-accuracy trade-off. 

\begin{figure*}
        \centering
         
     \subfloat[]{\includegraphics[width=.55\linewidth]{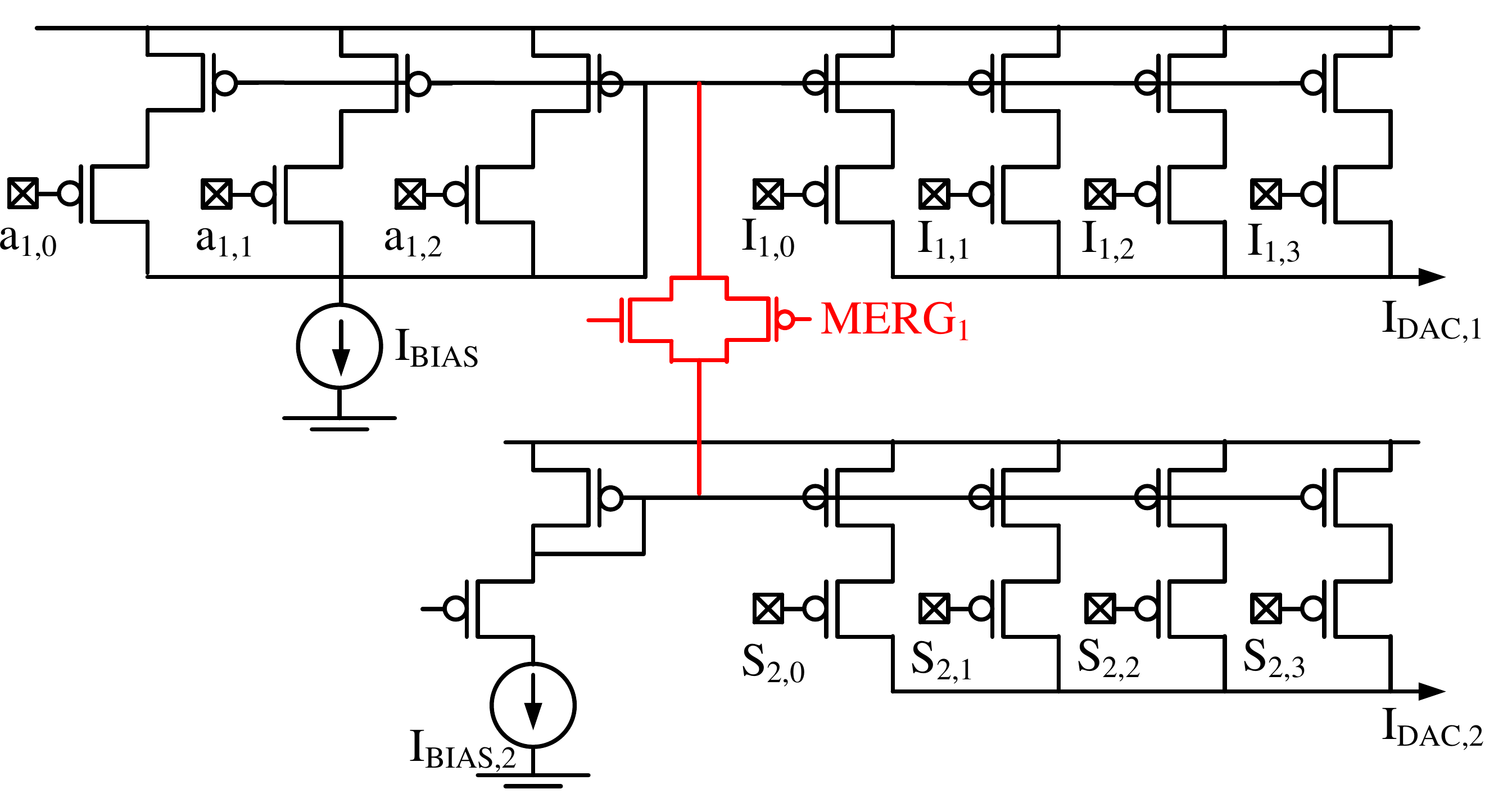}}  
       \subfloat[]{\includegraphics[width=.4\linewidth]{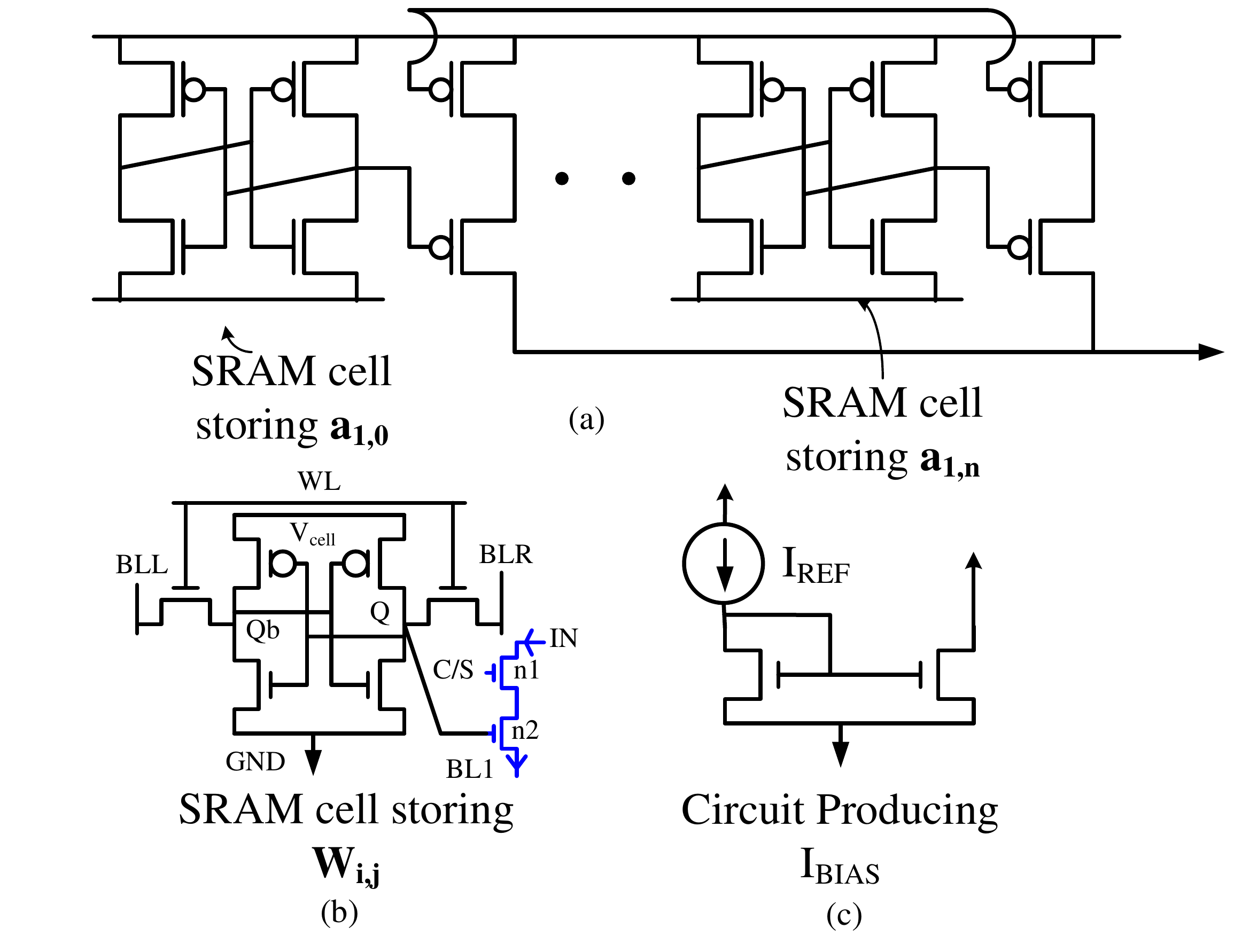}} 
         \caption{(a) Digital to analog converter (DAC) with support parameter $a_{i,j}$. (b) 8-T SRAM bitcell and current mirror which supply the $I_{BIAS}$ .}
         \label{fig:RD1}
    \end{figure*}
Fig. \ref{fig:SRAMB} shows the implementation of SRAM-BNN with bitcell array-based support parameters. The architecture also comprises current steering digital-to-analog converter (DAC), successive approximation analog-to-digital converter (ADC), row decoder, column multiplexer, and activation function processing unit in the peripherals. The input vector to a neuron layer $I$ is stored at the DAC operand buffer at the right in the figure. Input values are converted to the analog mode current $I_{DAC}$ using row digital-to-analog converter (DAC) array. SRAM cells in the array are designed using eight transistors and have an additional scalar product port [ Fig. \ref{fig:RD1}(b.b)]. SRAM cells operate as a current mode AND gate. If an SRAM cell stores ‘1’, it allows row DAC current $I_{DAC,j}$ to flow through its bit line. Otherwise, the current is blocked. Scalar product port of SRAM can be selected using column select signal(C/S) in the figure. Total column current in column $i$, therefore, follows the scalar product of $I \cdot w_i$, where $w_i$ is the binary weight vector stored in the column. The key features of the proposed implementation are: (i) The SRAM array operates in a column-wise parallel mode, which enables a high throughput processing. (ii) The 8-T SRAM decouples read/write operation with the computation of scalar products. As a result, the scalar product does not interfere with the typical operation of cells. 

Fig. \ref{fig:RD1} shows the DAC architecture with support parameters we use in the proposed design. We revised the current steering DAC as shown in Fig. \ref{fig:RD1}(a). $a_{i,j}$ are the support parameters stored in the SRAM cell [Fig. \ref{fig:RD1}(b.a)]. We chose SAR ADC for the digital conversion as it does not require large capacitance for low precision. Complete digital nature of SAR ADCs enables low power overhead and lower complexity. A nonlinear activation function is typically used in every layer of the fully connected neural network. In our work, we used the Relu activation function. The Relu activation function has become popular in recent years due to its simplicity. It also has the ability to enable fast training.

\section{Experimental Results and Discussions}

\begin{figure}[]
        \centering
        \subfloat[]{\includegraphics[width=0.18\linewidth]{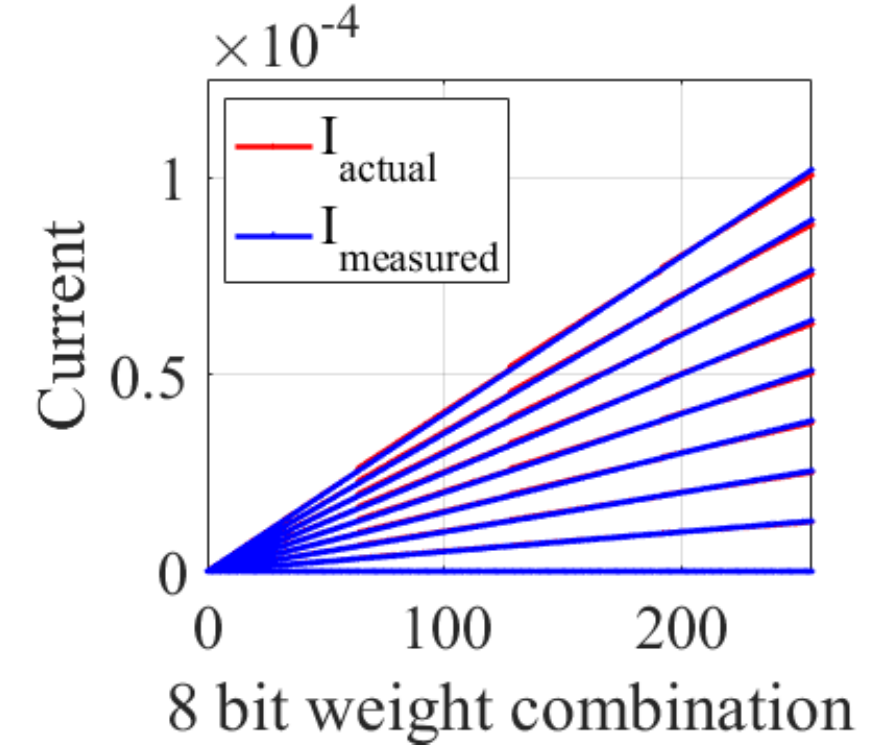}}
        \subfloat[]{\includegraphics[width=0.18\linewidth]{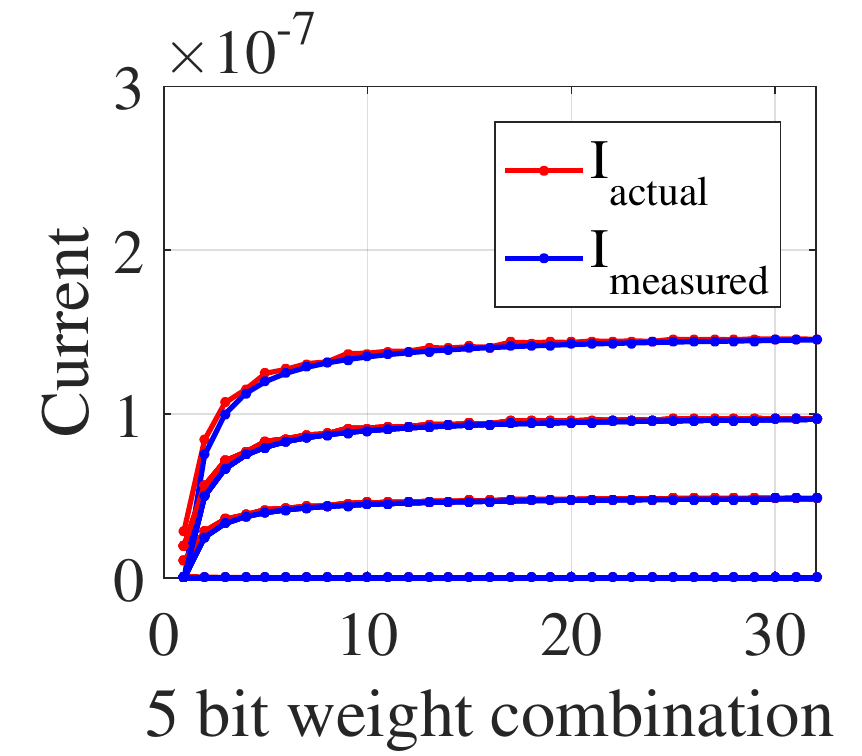}}
         \subfloat[]{\includegraphics[width=0.21\linewidth]{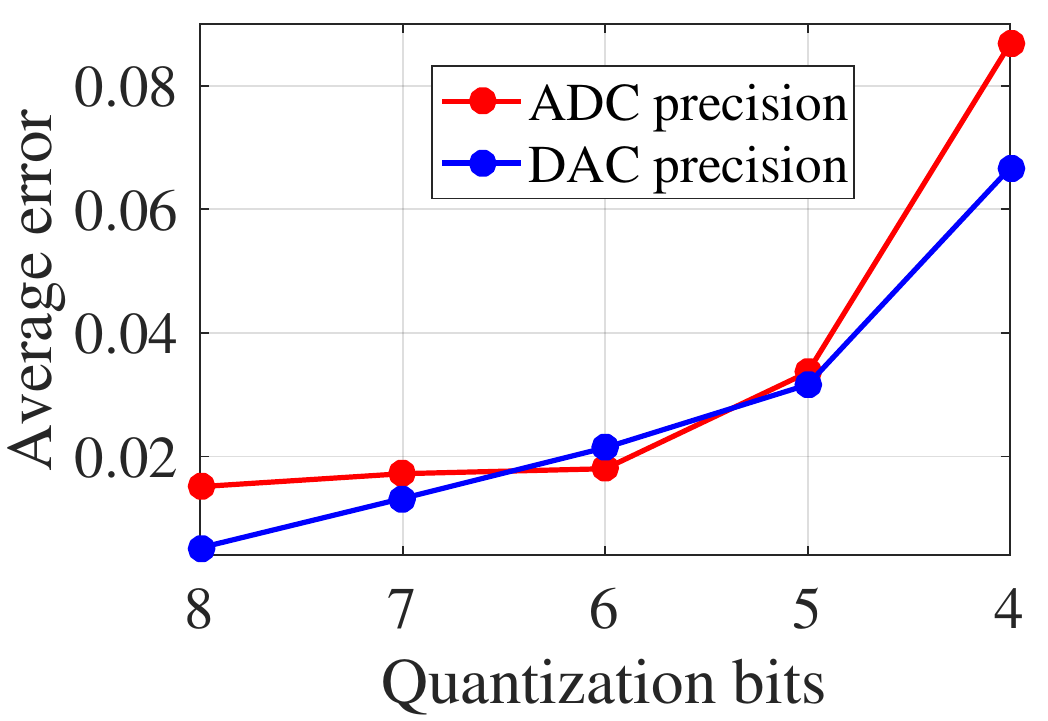}}
    \subfloat[]{\includegraphics[width=0.21\linewidth]{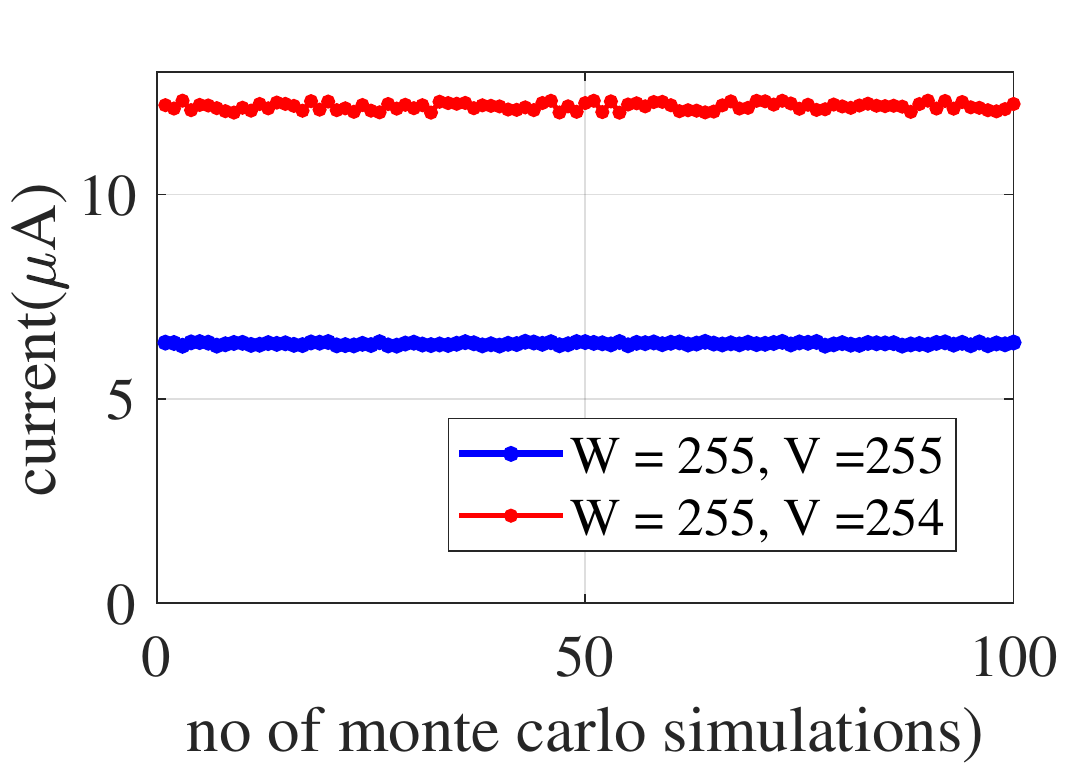}}
   \subfloat[]{\includegraphics[width=0.21\linewidth]{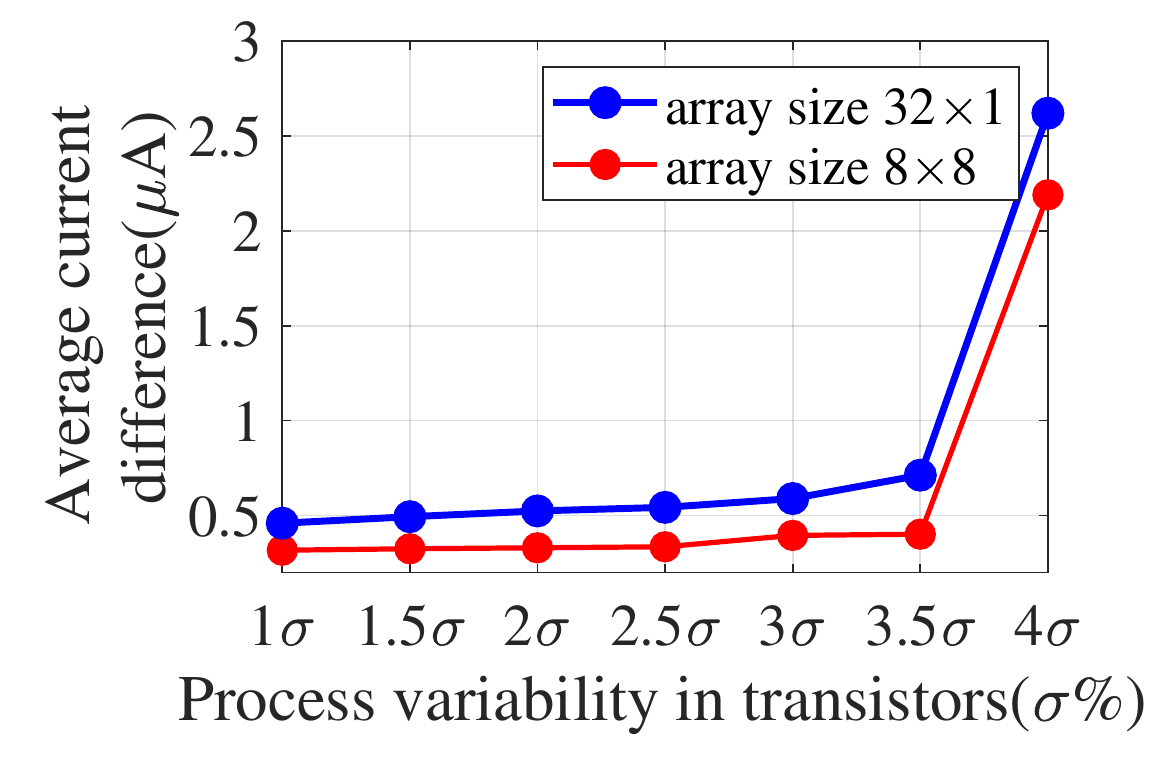}}
         \caption{(a) actual and experimental current difference for $8\times 1$ block without introducing any support parameters. (b) actual and experimental current for $32\times 1$ block with 5-bit support parameter. (c) accuracy of a  fully connected MLP for MNIST training data set with different  bit precision of the DAC and ADC. Effect of $V_{TH}$ variability in SRAM cell transistors to scalar product current. (d) Current for two different weight and feature combination at 30 mV processes variation in $V_{th}$ after 100 monte carlo simulation. (e) Scalar product(as current) at different level of processes variation.}
         \vspace{-1.3em} 
         \label{fig:Cur}
    \end{figure}

We simulated the functional analysis in Pytorch and MATLAB and the experimental analysis in HSPICE.  We initialize the weights $W_{i,j}$, support parameters $a_{i,j}$ and $b_{i,j}$ randomly in Pytorch and store them in SRAM array. We perform the computation of the scalar product of $I_{DAC}.W/a$  in HSPICE. A 5-bit current steering DAC, which takes the binarized feature vector as input, produces the current $I_{DAC}$. We read the product data from HSPICE data in Pytorch to train the network for 100 epochs. After training, we perform the inference of the test data set in Pytorch. We wrote a MATLAB script for HSPICE simulation. For current,power-performance, and process variation study, the simulations use 45nm technology node. Fig. \ref{fig:Cur}(a) shows the current analysis of an $8\times 1$ array. Here we swept the input value from 0-255 as the pixel values of the MNIST data set varies from 0-255 for different weight combination. From the figure, it is evident that the experimental current value matches the theoretical current value. Inclusion of support parameters reduces the current to $\sum I_{DAC}/a$ from $\sum I_{DAC}$, where $a$ is the support parameter. Fig. \ref{fig:Cur}(b) shows the current results with 5-bit support parameters.

Previously, the Neural Network models were designed to maximize the accuracy without considering complexity of implementation, which requires high energy cost. Reduction of the bit precision of the operands and operation is one way to reduce the implementation complexity and power consumption, which includes the conversion of floating-point to fixed-point, reducing the bitwidth, non-linear quantization, and weight sharing. In our work, all the weights stored in SRAM are 1-bit, and we varied bit precision of support parameter from 8-bit to 3-bit to analyze the accuracy results. We also varied the bit precision of the DAC and ADC in our network. We varied both the precision from 8-bit to 4-bit.  Fig. \ref{fig:Cur}(c) shows the average error value of the architecture for different bit precision of DAC and ADC. We simulated the architecture in bank by bank. We choose different bank sizes to analyze the power performance of our architecture. The architecture consumes power of 48.09 $\mu W$ for bank size $8\times 8$ and  156.04 $\mu W$ for bank size $64\times 64$.

While keeping the cell design to the simplest in previous work, the scalar product is sensitive to the threshold voltage ($V_{T H}$) variability in SRAM cells. Meanwhile, in our proposed design, the SRAM cells either act as a current buffer or block the input current so that the variability in cell transistors has a minimal impact on the accuracy of a scalar product.  Fig. \ref{fig:Cur}(d-e) shows the effect of process variation. We conducted 100 monte Carlo simulations with weight matrix W = 255 (32 bits), feature matrix V = 255 and 254 for a process variation of 30 mV. From the figure, it is evident that our architecture is robust to processes variation.  We also simulated the architecture for various processes variability (Fig. \ref{fig:Cur}(e)). 
\section{Conclusion}
In this paper, we introduce a bitcell array-based supported-BinaryNet, which achieves higher prediction accuracy than the SRAM-based binarized neural network (SRAM-BNN) by enhancing the training weight space of SRAM-BNN while requiring minimal overheads to a typical design. Compared to a typical SRAM-BNN, our approach suffers from power and latency overheads. We propose a dynamic drop out of a part of the support parameters to reduce the power overheads. Our proposed architecture is able to tolerate lower precision of transistors and other component variabilities.

\section*{References}
\small





[1] D. Keitel-Schulz and N. Wehn, “Embedded dram development: Technology, physical design, and application issues,”, May 2001.

[2]  Y. Chen, T. Luo, S. Liu, S. Zhang, L. He, J. Wang, L. Li, T. Chen, Z. Xu, N. Sun, and O. Temam, “Dadiannao: A machine-learning supercomputer,” 2014.

[3] Jintao Zhang, Zhuo Wang, and N. Verma, “A machine-learning classifier implemented in a standard 6t sram array,” in 2016 VLSI-Circuits, June 2016.

[4] P. Chi, S. Li, C. Xu, T. Zhang, J. Zhao, Y. Liu, Y. Wang, and Y. Xie, “Prime: A novel processing-in-memory architecture for neural network computation in reram-based main memory,” Jun. 2016.

[5] I. Hubara, M. Courbariaux, D. Soudry, R. El-Yaniv, and Y. Bengio, “Binarized neural networks,” , 2016.

[6] D. P. Kingma and J. Ba, “Adam: A Method for Stochastic Optimization,” 2014.

\end{document}